\documentclass[10pt,conference]{IEEEtran}

\usepackage{cite}
\usepackage{amsmath,amssymb,amsfonts}
\usepackage{algorithmic}
\usepackage{graphicx}
\usepackage{textcomp}
\usepackage{xcolor}
\usepackage{booktabs}
\usepackage{url}
\usepackage{balance}

\usepackage[table]{xcolor}
\usepackage{colortbl}
\definecolor{color1}{HTML}{789ECB}
\definecolor{color2}{HTML}{A0C0DD}
\definecolor{color3}{HTML}{F6B7AD}
\definecolor{color4}{HTML}{E97A75}

\newcommand{\hm}[1]{%
  \begingroup
  \color{black}%
  \ifdim#1pt<15pt
      \cellcolor{color4!50}
  \else\ifdim#1pt<30pt
      \cellcolor{color4!30}
  \else\ifdim#1pt<45pt
      \cellcolor{color3!30}
  \else\ifdim#1pt<60pt
      \cellcolor{color3!10}
  \else\ifdim#1pt<75pt
      \cellcolor{color2!10}
  \else\ifdim#1pt<90pt
      \cellcolor{color2!30}
  \else\ifdim#1pt<95pt
      \cellcolor{color1!30}
  \else
      \cellcolor{color1!50}
  \fi\fi\fi\fi\fi\fi\fi
  #1%
  \endgroup
}

\begin{document}

\title{Beyond the Answer Key: Robustness Evaluation of Large Language Models for Step-Level Mathematical Verification}

\author{
\IEEEauthorblockN{Fateme Mazdarani}
\IEEEauthorblockA{
School of Computing\\
Clemson University\\
Clemson, SC, USA\\
fmazdar@clemson.edu
}
\and
\IEEEauthorblockN{Carlos Toxtli}
\IEEEauthorblockA{
School of Computing\\
Clemson University\\
Clemson, SC, USA\\
ctoxtli@clemson.edu
}
}

\maketitle

\begin{abstract}
Large language models (LLMs) are increasingly used as graders, verifiers, and process auditors, but most mathematical evaluations still emphasize final-answer accuracy. This can obscure whether a model can verify a non-canonical but valid solution trace. We introduce a controlled linear-equation benchmark for evaluating LLMs in the evaluator role. Each instance asks the model to judge final-answer correctness, step-level trace correctness, and the first incorrect step. Our evaluation of state-of-the-art open LLMs reveals a significant robustness gap: models that accurately evaluate canonical solutions often fail when presented with perturbed but logically equivalent variants. Across GPT-OSS 20B, Qwen3-14B, and Phi-4-Reasoning, base models perform well on canonical traces but degrade substantially on perturbed traces, especially for error localization. On valid perturbed traces, base-model false-rejection rates reach 75.6--85.3\%, showing strong sensitivity to canonical solution form. Supervised fine-tuning, distillation, and test-time compute improve robustness in some settings, but gains are model dependent and can trade off against canonical performance. The results show that reliable process-level verification remains challenging, and evaluator robustness should be measured separately from solver accuracy, even in a simple algebraic domain with exact ground truth.
\end{abstract}


\section{Introduction}
Large Language Models (LLMs) have demonstrated remarkable capabilities in supporting complex, multi-step tasks across education, software engineering, and scientific research. In correctness-critical applications, the validity of the final result depends entirely on the integrity of the underlying process, and a single error in the process can invalidate the whole result. Therefore, LLMs, when functioning as evaluators, must be capable of distinguishing between valid alternative methods and genuine logical failures. Automated evaluators must be accurate and robust since users may trust explained outputs without necessarily understanding their limitations~\cite{BhandariPardosEDM2025Autograder,lightman2023lets,mazdarani2026}.

\begin{figure}[t]
    \centering
    \includegraphics[width=\linewidth]{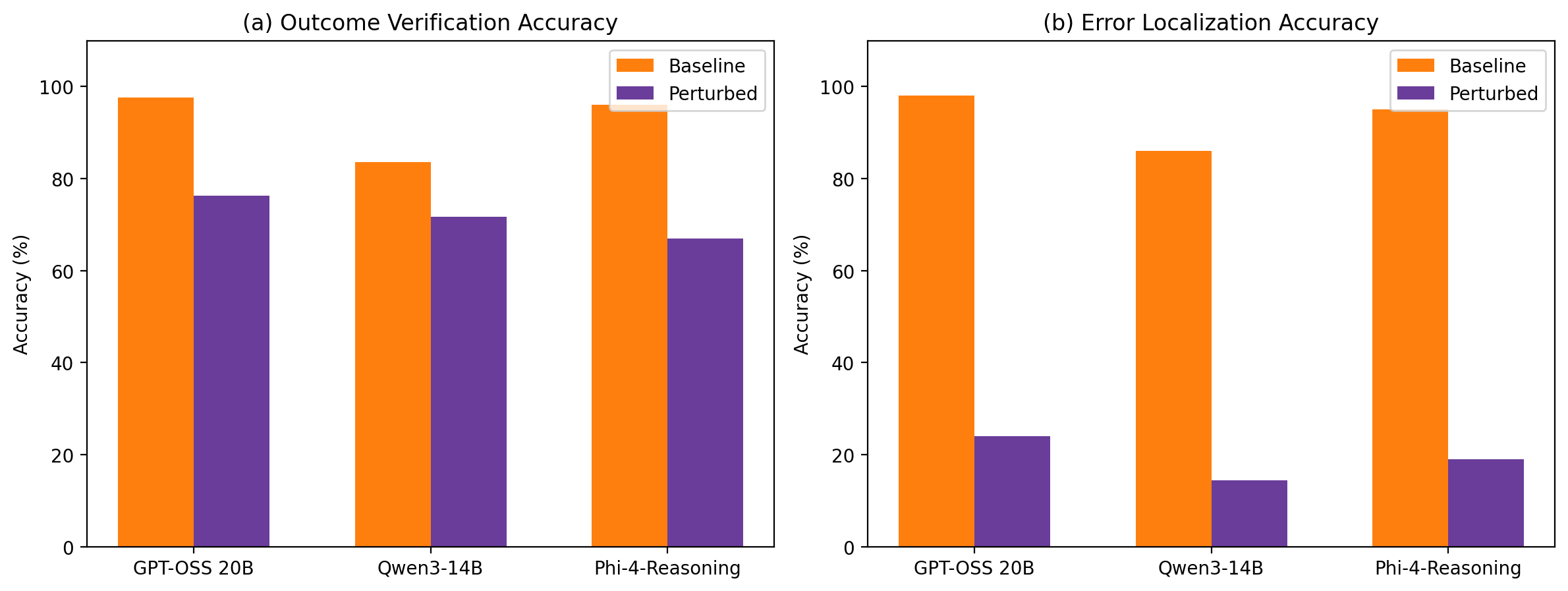}
    \caption{Robustness gap for base models. Final-answer evaluation is affected by solution perturbations, but the larger degradation occurs in first-error localization, indicating that the models often fail to separate valid non-canonical transformations from true errors.}
    \label{fig:robustness_gap}
\end{figure}

In mathematical contexts, LLMs are not only used to solve problems  but also to evaluate solutions produced by humans, tools, or other models. In such evaluator settings, final-answer correctness is insufficient: a solution with the right answer may contain an invalid step, and a correct solution may use a non-standard path that should not be penalized. 
Most mathematical LLM benchmarks emphasize the solver role and report outcome-based metrics such as final-answer accuracy on datasets including GSM8K~\cite{DBLP:journals/corr/abs-2110-14168} and MATH~\cite{DBLP:journals/corr/abs-2103-03874}. This focus can hide a different failure mode: a model may grade canonical worked solutions accurately while rejecting an alternative trace that is mathematically sound but unfamiliar in form. We call this failure mode \emph{template sensitivity}: dependence on a familiar solution template rather than on step-level validity. As shown in \figurename~\ref{fig:robustness_gap}, this sensitivity is especially severe for first-error localization, where several models that appear reliable on canonical traces collapse under solution-level perturbations.

We study this problem in linear equations of the form $ax+b=cx+d$. This domain is intentionally narrow: it gives exact ground truth, controllable trace generation, and unambiguous error localization. We do not claim that linear equations capture all mathematical or scientific verification tasks. Instead, we use them as a diagnostic setting in which failures cannot be attributed to ambiguity in the problem statement or to subjective grading criteria.
Our contributions are as follows:
\begin{itemize}
    \item \textbf{Controlled trace-perturbation benchmark:} We hold each equation fixed while varying perturbed traces, isolating evaluator robustness with exact labels for correct, incorrect, and tricky solutions.
    \item \textbf{Three-part evaluation protocol:} We evaluate final-answer correctness ($Q_1$), step-level correctness ($Q_2$), and first-error localization ($Q_3$), enabling separate measurement of answer checking, process checking, and diagnostic feedback.
    \item \textbf{Empirical and adaptation analysis:} We evaluate leading open models GPT-OSS 20B~\cite{OpenAI2025GPTOSS}, Qwen3-14B~\cite{Alibaba2025Qwen3}, and Phi-4-Reasoning~\cite{Microsoft2025Phi4}, quantifying the performance gap between baseline and perturbed alternatives, and identifying common failure modes; then test supervised fine-tuning, Gemini-enhanced distillation, and test-time compute.
\end{itemize}

\section{Related Work}
\subsection{Mathematical Reasoning Benchmarks and Verification}
A large body of work evaluates LLMs on mathematical reasoning by measuring solving performance on controlled problem sets such as MATH~\cite{DBLP:journals/corr/abs-2103-03874}. Such benchmarks primarily assess LLMs as problem solvers rather than evaluators. Related verifier work trains process-based reward models or outcome reward models for scoring generated reasoning traces~\cite{lightman2023lets,wang2024mathshepherd}. More recent work studies step-level verifier training, verifier-guided inference, and formally verified process labels~\cite{chang2025hybridtts,kamoi2025formalverifiers,li2025verifybench}. Our work complements these efforts by focusing on the grading task itself: whether a model assigns consistent judgments to worked solutions that are procedurally different but algebraically admissible.

\subsection{Robustness under Perturbations}
Recent studies show that LLM mathematical reasoning can be fragile under perturbations to problem statements, irrelevant context, symbolic forms, or numerical values~\cite{li2024gsm,mirzadeh2024gsmsymbolic,huang2025mathperturb,chatziveroglou2025exploring,numericalsensitivity2025}. These studies mainly examine the solver setting. We instead perturb the \emph{solution trace} while holding the original equation fixed. This isolates evaluator robustness: the model is not asked to solve a new problem, but to judge whether a supplied derivation is valid.

\subsection{Evaluating Worked Solutions}
Recent work has explored using LLMs as autograders for worked solutions in mathematics by comparing LLM-assigned correctness labels against human grading~\cite{BhandariPardosEDM2025Autograder}. Other step-by-step evaluation settings, though not focused on math grading specifically, also emphasize measuring intermediate reasoning behavior rather than only final outputs~\cite{chen2024teval}. Such studies demonstrate the promise of automated evaluation, but they typically do not systematically control solution-trace variability. Our benchmark fills this gap by generating paired canonical and perturbed traces with exact labels for the first invalid step.

\section{Methodology}
\subsection{Dataset Generation and Taxonomy}
\begin{figure}[t]
    \centering
    \includegraphics[width=\linewidth]{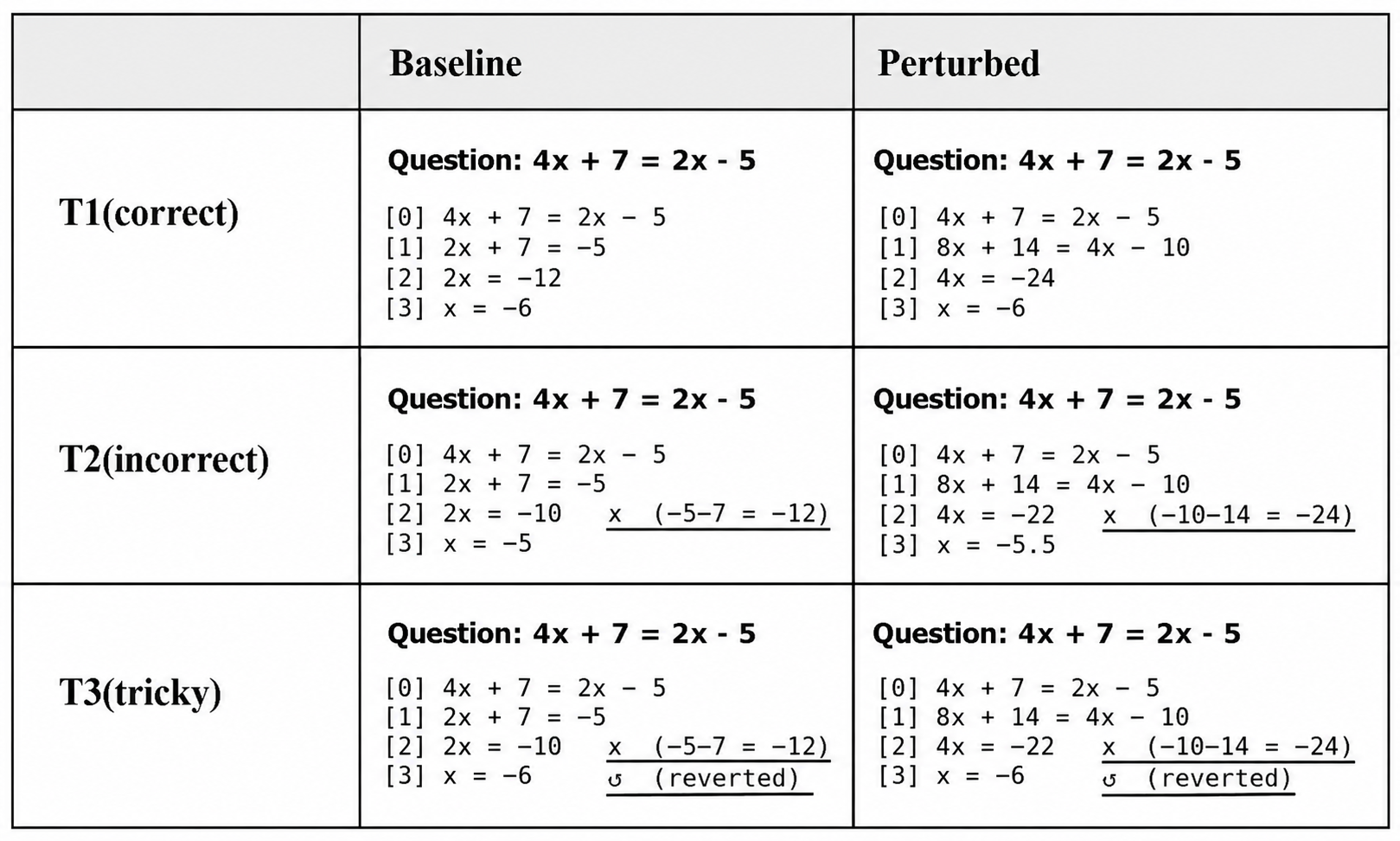}
    \caption{Benchmark taxonomy. The same equation can appear with a canonical or perturbed trace and as T1, T2, or T3. This design separates final answer from trace correctness; the evaluation prompt does not contain underlined text.}
    \label{fig:taxonomy_table_baseline_perturbed}
\end{figure}

We generate equations of the form $ax+b=cx+d$ with a unique rational solution. A solving engine then emits worked traces under three trace types:
\begin{itemize}
    \item \textbf{T1 (Correct):} Every step maintains mathematical equivalence under the trace semantics in Section~\ref{sec:semantics}, and the final answer is correct. T1 is used to measure false rejection of valid reasoning.
    \item \textbf{T2 (Incorrect):} A single arithmetic or algebraic error is injected at a known step and propagated to an incorrect final answer. T2 tests whether the evaluator detects and localizes the true first error.
    \item \textbf{T3 (Tricky):} A trace contains an invalid intermediate step, but the final answer is manually reverted to the correct value. T3 separates final-answer checking from process checking.
\end{itemize}
The benchmark taxonomy is illustrated in \figurename~\ref{fig:taxonomy_table_baseline_perturbed}.

\subsection{Reproducibility Details}
Equation instances are generated by sampling integer coefficients $a,b,c,d$ from $[-10,10]$ and enforcing $a\neq c$, so the solution $x=(d-b)/(a-c)$ is unique. We reject degenerate equations, empty transformations, and traces whose verification obligations cannot be discharged. Train and test splits are instance-disjoint: we hash the tuple $(a,b,c,d)$ before trace generation so that no equation appears in both training and evaluation.

Step indexing is 0-based. Step $0$ is the original equation string. The first-error label $Q_3$ is the index of the first invalid step, and $Q_3=-1$ when no invalid step exists. All outputs are requested as strict JSON. A malformed response is recorded as a parse error and counted as incorrect for the affected metrics.

\subsection{Trace Semantics and Validation}\label{sec:semantics}
A central concern in this benchmark is that not every familiar algebraic manipulation is an equivalence over the real line. For example, multiplying both sides by an expression that may be zero or squaring both sides can introduce extraneous candidates. We therefore label traces using a checker semantics rather than treating every syntactic transformation as a bidirectional equivalence.

Let $S_i(x)$ denote the solution set represented by step $i$, and let $O_i$ denote any active obligation, such as verifying a candidate or recording a nonzero denominator. A step is accepted as \emph{strictly equivalent} when $S_i(x)\Leftrightarrow S_{i+1}(x)$ over the active domain. This category includes adding the same expression to both sides, moving terms across the equality sign with sign change, and multiplying or dividing by a known nonzero constant. A step is accepted as \emph{candidate-generating} when it preserves all true solutions but may introduce extra candidates, i.e., $S_i(x)\Rightarrow S_{i+1}(x)$, provided that the trace later verifies every candidate against the original equation and discards extraneous roots. Multiplication by a variable term and squaring both sides are handled in this second category. A step is labeled invalid when it contains an arithmetic error, drops a valid candidate, introduces an unsupported restriction, or leaves an obligation unresolved.
Accordingly, $Q_2=1$ if and only if every transition is either strictly equivalent or candidate-generating and $Q_3$ is the first transition where this condition fails. A deterministic exact-arithmetic trace checker is used to generate and validate all gold labels.

\subsection{Perturbation Families}
\begin{figure}[t]
    \centering
    \includegraphics[width=\linewidth]{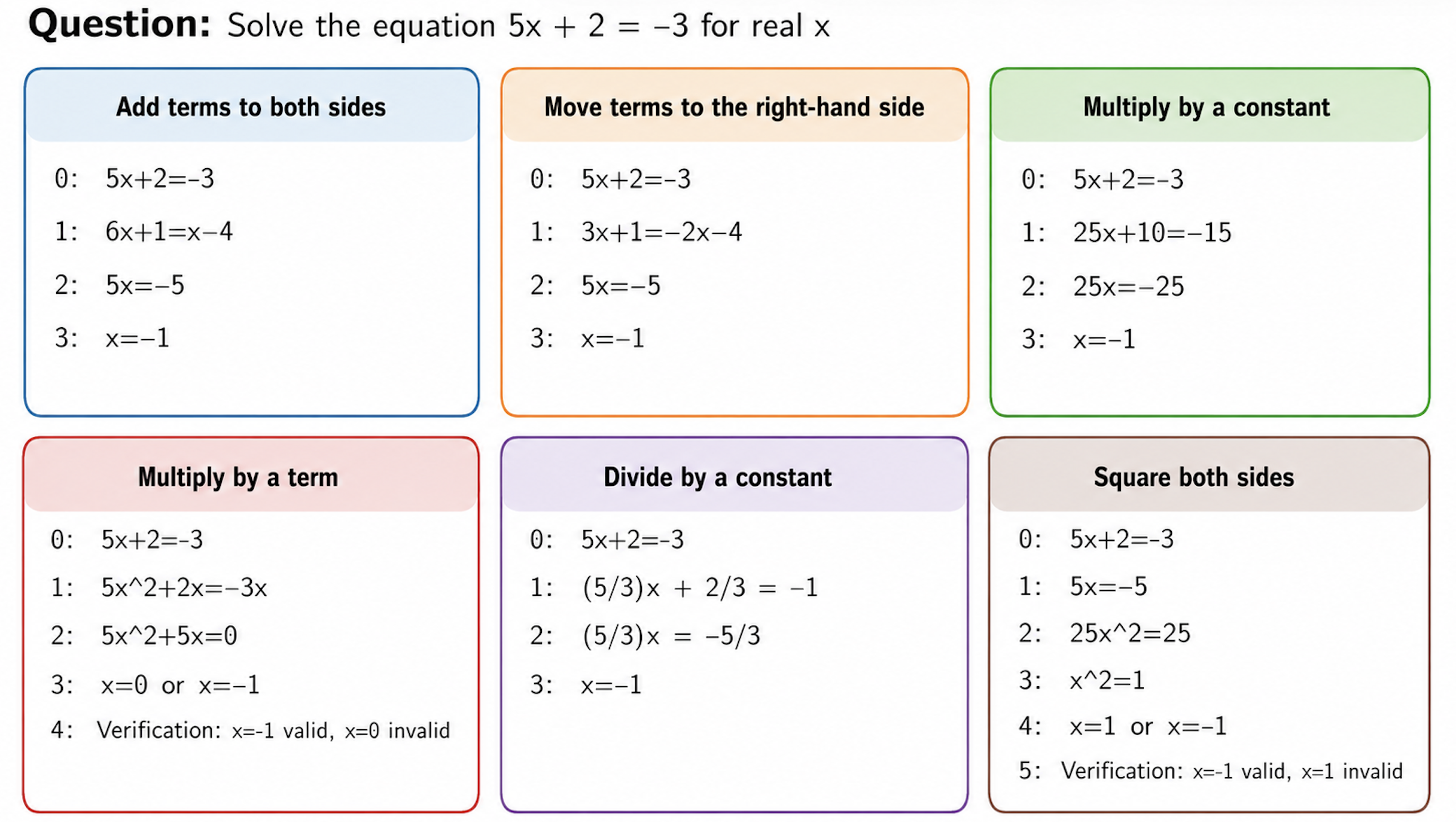}
    \caption{Perturbation families. The first four panels are equivalence-preserving in the generated domain. Multiplication by a term and squaring both sides are candidate-generating transformations; they are accepted only when all resulting candidates are explicitly checked against the original equation.}
    \label{fig:perturbation_types}
\end{figure}

A core feature of this benchmark is the inclusion of perturbed solutions through systematic injection of mathematical perturbations--valid but unconventional transformations that deviate from the shortest path to a solution. 
The benchmark includes six perturbation families, shown in \figurename~\ref{fig:perturbation_types}. Four families are strictly equivalence-preserving in our domain: adding terms to both sides, moving terms to the right-hand side, multiplying by a nonzero constant, and dividing by a nonzero constant. Two families are obligation-generating: multiplying by an expression involving $x$ and squaring both sides. For these cases, the generated trace explicitly verifies the resulting candidate set by substitution into the original equation. 
These controlled stress tests assess whether models distinguish valid alternative paths from errors; they are not intended to represent the full distribution of student work.

\subsection{Evaluation Framework}
For each sample, the model receives the original equation and the complete worked trace. It must return a JSON object with three fields: (i) $Q_1$, a Boolean indicating whether the final answer satisfies the original equation; (ii) $Q_2$, a Boolean indicating whether the full trace is mathematically valid; and (iii) $Q_3$, the integer index of the first invalid step or $-1$ if no invalid step exists. This format separates outcome verification from process verification and diagnostic localization.

We report accuracy for each query overall and stratified by T1/T2/T3. For T1 traces, $1-\mathrm{Acc}(Q_2)$ is the false-rejection rate:
the evaluator rejects a valid trace. For T2 and T3 traces, a high $Q_2$ error rate indicates false acceptance of an invalid process. Localization accuracy is intentionally strict and requires exact agreement with the first-error index.

\subsection{Model Selection and Adaptation}
We evaluate three open-weight reasoning-oriented models: GPT-OSS 20B~\cite{OpenAI2025GPTOSS}, Qwen3-14B~\cite{Alibaba2025Qwen3}, and Phi-4-Reasoning~\cite{Microsoft2025Phi4}. We selected these models after preliminary screening to avoid conflating evaluator robustness with a basic inability to follow mathematical notation. All experiments are implemented using the Hugging Face \texttt{Transformers} and \texttt{PEFT} libraries in \texttt{bf16} precision with gradient checkpointing enabled. Non-TTC evaluations use deterministic decoding; TTC uses seven independent samples at temperature $0.3$ followed by majority vote for each query.

\begin{table*}[t]
\caption{Test accuracy (\%) for GPT-OSS-20B-based models}
\label{tab:table1_gptoss}
\centering
\scriptsize
\setlength{\tabcolsep}{2.5pt}
\renewcommand{\arraystretch}{1.2}

\begin{tabular}{lcccc cccc cccc}
\toprule
\textbf{Model} &
\multicolumn{4}{c}{\textbf{Outcome Accuracy ($Q_1$)}} &
\multicolumn{4}{c}{\textbf{Step-Level Accuracy ($Q_2$)}} &
\multicolumn{4}{c}{\textbf{Localization Accuracy ($Q_3$)}} \\
\cmidrule(lr){2-5}\cmidrule(lr){6-9}\cmidrule(lr){10-13}
& \textbf{Overall} & \textbf{T1: Correct} & \textbf{T2: Incorrect} & \textbf{T3: Tricky}
& \textbf{Overall} & \textbf{T1: Correct} & \textbf{T2: Incorrect} & \textbf{T3: Tricky}
& \textbf{Overall} & \textbf{T1: Correct} & \textbf{T2: Incorrect} & \textbf{T3: Tricky} \\
\midrule

\multicolumn{13}{l}{\textbf{Baseline}} \\
Base
& \hm{97.69} & \hm{98.33} & \hm{98.06} & \hm{96.67}
& \hm{98.24} & \hm{98.33} & \hm{98.06} & \hm{98.33}
& \hm{98.06} & \hm{98.33} & \hm{97.78} & \hm{98.06} \\
Base + TTC
& \hm{85.83} & \hm{100} & \hm{99.17} & \hm{58.33}
& \hm{99.35} & \hm{100} & \hm{98.89} & \hm{99.17}
& \hm{80.09} & \hm{100} & \hm{77.50} & \hm{61.94} \\
FT
& \hm{81.48} & \hm{96.39} & \hm{61.67} & \hm{86.39} 
& \hm{82.13} & \hm{75.00} & \hm{88.89} & \hm{82.50} 
& \hm{74.17} & \hm{74.72} & \hm{70.56} & \hm{77.22} \\
FT + TTC
& \hm{68.98} & \hm{97.50} & \hm{36.67} & \hm{72.78}
& \hm{69.72} & \hm{93.89} & \hm{51.11} & \hm{64.17}
& \hm{66.39} & \hm{96.11} & \hm{41.11} & \hm{61.94} \\
GED-FT
& \hm{77.31} & \hm{97.78} & \hm{85.83} & \hm{48.33}
& \hm{89.63} & \hm{98.06} & \hm{86.11} & \hm{84.72}
& \hm{85.74} & \hm{96.94} & \hm{80.28} & \hm{80.00} \\
GED-FT + TTC
& \hm{84.63} & \hm{100} & \hm{61.94} & \hm{92.22}
& \hm{76.85} & \hm{100} & \hm{72.78} & \hm{57.78}
& \hm{74.72} & \hm{100} & \hm{70.56} & \hm{52.78} \\
\midrule

\multicolumn{13}{l}{\textbf{Perturbed}} \\
Base
& \hm{76.30} & \hm{72.50} & \hm{88.61} & \hm{67.78}
& \hm{54.06} & \hm{24.44} & \hm{71.72} & \hm{72.00}
& \hm{24.07} & \hm{24.72} & \hm{23.06} & \hm{24.44}\\
Base + TTC
& \hm{32.04} & \hm{38.06} & \hm{44.72} & \hm{13.33}
& \hm{31.11} & \hm{18.61} & \hm{45.83} & \hm{28.89}
& \hm{35.07} & \hm{95.83} & \hm{6.94} & \hm{3.00} \\
FT
& \hm{73.33} & \hm{95.00} & \hm{31.39} & \hm{93.61}
& \hm{69.35} & \hm{23.06} & \hm{92.78} & \hm{92.22}
& \hm{42.87} & \hm{23.33} & \hm{47.22} & \hm{58.06} \\
FT + TTC
& \hm{63.89} & \hm{83.61} & \hm{35.00} & \hm{73.06}
& \hm{47.59} & \hm{74.72} & \hm{40.28} & \hm{27.78}
& \hm{43.89} & \hm{86.94} & \hm{27.50} & \hm{16.94} \\
GED-FT
& \hm{84.35} & \hm{95.83} & \hm{73.61} & \hm{83.61}
& \hm{76.94} & \hm{94.72} & \hm{79.44} & \hm{56.67}
& \hm{58.98} & \hm{93.61} & \hm{45.00} & \hm{38.33} \\
GED-FT + TTC
& \hm{83.70} & \hm{99.44} & \hm{56.11} & \hm{96.56}
& \hm{66.20} & \hm{98.33} & \hm{68.33} & \hm{31.06}
& \hm{56.50} & \hm{100} & \hm{42.50} & \hm{25.06} \\
\bottomrule
\end{tabular}
\end{table*}

\begin{itemize}
\item \textbf{Fine-Tuning (FT):} We perform parameter-efficient fine-tuning with LoRA ($\mathrm{lr}=2\times10^{-4}$, $r\in\{8,16\}$, $\alpha\in\{16,32\}$) for 3 epochs with batch size 16. T1 samples teach acceptance of valid non-canonical traces; T2 and T3 samples teach rejection and localization of invalid transitions.
\item \textbf{Gemini-Enhanced Distillation (GED-FT):} We refine the programmatic explanations using Gemini 3 Flash~\cite{Google2025Gemini3}. To prevent teacher errors from corrupting the labels, we retain only samples for which the teacher explanation preserves the ground-truth $(Q_1,Q_2,Q_3)$ labels. This filtering yields 4,000 training samples. The goal is to test whether richer natural-language rationales improve verifier behavior beyond label-only supervision.
\item \textbf{Test-Time Compute (TTC):} We aggregate $k=7$ sampled responses by majority vote for each output field. TTC tests whether additional reasoning samples correct isolated mistakes or instead amplify systematic template sensitivity.
\end{itemize}

\subsection{Data Composition and Partitioning}
The training set contains 27,000 samples: 9,000 canonical traces, evenly split across T1/T2/T3, and 18,000 perturbed traces, evenly split across trace type and perturbation family. Evaluation uses two held-out sets of 1,080 samples each: one canonical and one perturbed. Each evaluation set has 360 samples per trace type; the perturbed set is balanced across perturbation families. For each sample, the released metadata records the equation tuple, trace type, perturbation family, obligation log, gold labels, and checker explanation.
The benchmark, code, and experimental configurations are available at
\url{https://github.com/mazdarani/beyond-answer-key}.

\section{Results and Discussion}
\subsection{Base-Model Robustness Gap}
Tables~\ref{tab:table1_gptoss}, \ref{tab:table2_Qwen}, and \ref{tab:table3_phi4} show that base models perform well on canonical traces but degrade sharply under perturbation. The largest degradation is in $Q_3$, confirming that first-error localization is substantially harder than final-answer checking. The drop is not merely a formatting artifact: for valid perturbed T1 traces, the base models falsely reject correct reasoning at high rates according to $Q_2$: 75.56\% for GPT-OSS 20B, 83.89\% for Qwen3-14B, and 85.28\% for Phi-4-Reasoning. These rates clearly quantify template sensitivity.

\begin{table*}[t]
\caption{Test accuracy (\%) for Qwen3-14b-based models}
\label{tab:table2_Qwen}
\centering
\scriptsize
\setlength{\tabcolsep}{2.5pt}
\renewcommand{\arraystretch}{1.2}

\begin{tabular}{lcccc cccc cccc}
\toprule
\textbf{Model} &
\multicolumn{4}{c}{\textbf{Outcome Accuracy ($Q_1$)}} &
\multicolumn{4}{c}{\textbf{Step-Level Accuracy ($Q_2$)}} &
\multicolumn{4}{c}{\textbf{Localization Accuracy ($Q_3$)}} \\
\cmidrule(lr){2-5}\cmidrule(lr){6-9}\cmidrule(lr){10-13}
& \textbf{Overall} & \textbf{T1: Correct} & \textbf{T2: Incorrect} & \textbf{T3: Tricky}
& \textbf{Overall} & \textbf{T1: Correct} & \textbf{T2: Incorrect} & \textbf{T3: Tricky}
& \textbf{Overall} & \textbf{T1: Correct} & \textbf{T2: Incorrect} & \textbf{T3: Tricky} \\
\midrule

\multicolumn{13}{l}{\textbf{Baseline}} \\
Base
& \hm{83.61} & \hm{96.94} & \hm{82.50} & \hm{71.39}
& \hm{86.48} & \hm{93.89} & \hm{82.50} & \hm{83.06}
& \hm{86.02} & \hm{93.89} & \hm{82.22} & \hm{81.94} \\
Base + TTC
& \hm{86.85} & \hm{100} & \hm{99.17} & \hm{61.39}
& \hm{99.63} & \hm{100} & \hm{98.89} & \hm{100}
& \hm{92.41} & \hm{100} & \hm{90.83} & \hm{86.39} \\
FT
& \hm{75.65} & \hm{55.56} & \hm{75.56} & \hm{95.83}
& \hm{78.33} & \hm{49.17} & \hm{89.17} & \hm{96.67}
& \hm{77.87} & \hm{49.17} & \hm{87.78} & \hm{96.67} \\
FT + TTC
& \hm{87.31} & \hm{80.83} & \hm{84.44} & \hm{96.67}
& \hm{93.70} & \hm{83.33} & \hm{98.61} & \hm{99.17}
& \hm{93.61} & \hm{85.00} & \hm{97.78} & \hm{97.78} \\
GED-FT
& \hm{71.94} & \hm{86.67} & \hm{43.61} & \hm{85.56}
& \hm{56.30} & \hm{86.67} & \hm{48.61} & \hm{33.61}
& \hm{52.50} & \hm{86.67} & \hm{40.28} & \hm{30.56} \\
GED-FT + TTC
& \hm{73.33} & \hm{100} & \hm{23.61} & \hm{96.39}
& \hm{53.70} & \hm{100} & \hm{47.78} & \hm{13.33}
& \hm{51.28} & \hm{100} & \hm{26.67} & \hm{27.17} \\
\midrule

\multicolumn{13}{l}{\textbf{Perturbed}} \\
Base 
& \hm{71.67} & \hm{90.28} & \hm{61.67} & \hm{63.06}
& \hm{52.69} & \hm{16.11} & \hm{68.06} & \hm{73.89}
& \hm{14.35} & \hm{16.11} & \hm{17.78} & \hm{9.17} \\
Base + TTC
& \hm{34.35} & \hm{28.06} & \hm{71.94} & \hm{3.06}
& \hm{53.61} & \hm{18.33} & \hm{72.50} & \hm{70.00}
& \hm{30.09} & \hm{80.00} & \hm{9.17} & \hm{0.00} \\
FT
& \hm{76.48} & \hm{70.83} & \hm{68.89} & \hm{89.72}
& \hm{77.87} & \hm{57.50} & \hm{88.06} & \hm{88.06}
& \hm{44.91} & \hm{57.50} & \hm{45.83} & \hm{31.39}
\\
FT + TTC
& \hm{78.24} & \hm{83.33} & \hm{81.39} & \hm{70.00}
& \hm{84.54} & \hm{73.33} & \hm{91.11} & \hm{89.17}
& \hm{81.67} & \hm{81.67} & \hm{83.33} & \hm{80.00} \\
GED-FT
& \hm{59.35} & \hm{87.78} & \hm{18.06} & \hm{72.22}
& \hm{41.85} & \hm{85.56} & \hm{29.44} & \hm{10.56}
& \hm{36.57} & \hm{85.56} & \hm{17.78} & \hm{6.39}  \\
GED-FT + TTC
& \hm{74.91} & \hm{98.61} & \hm{32.78} & \hm{93.33}
& \hm{56.57} & \hm{90.28} & \hm{57.78} & \hm{21.67}
& \hm{44.91} & \hm{88.89} & \hm{33.61} & \hm{12.22} \\
\bottomrule
\end{tabular}
\end{table*}

\begin{table*}[t]
\caption{Test accuracy (\%) for Phi-4-Reasoning-based models}
\label{tab:table3_phi4}
\centering
\scriptsize
\setlength{\tabcolsep}{2.5pt}
\renewcommand{\arraystretch}{1.2}

\begin{tabular}{lcccc cccc cccc}
\toprule
\textbf{Model} &
\multicolumn{4}{c}{\textbf{Outcome Accuracy ($Q_1$)}} &
\multicolumn{4}{c}{\textbf{Step-Level Accuracy ($Q_2$)}} &
\multicolumn{4}{c}{\textbf{Localization Accuracy ($Q_3$)}} \\
\cmidrule(lr){2-5}\cmidrule(lr){6-9}\cmidrule(lr){10-13}
& \textbf{Overall} & \textbf{T1: Correct} & \textbf{T2: Incorrect} & \textbf{T3: Tricky}
& \textbf{Overall} & \textbf{T1: Correct} & \textbf{T2: Incorrect} & \textbf{T3: Tricky}
& \textbf{Overall} & \textbf{T1: Correct} & \textbf{T2: Incorrect} & \textbf{T3: Tricky} \\
\midrule

\multicolumn{13}{l}{\textbf{Baseline}} \\
Base
& \hm{96.48} & \hm{92.22} & \hm{97.78} & \hm{99.44}
& \hm{96.11} & \hm{92.22} & \hm{97.78} & \hm{98.33}
& \hm{95.93} & \hm{92.22} & \hm{97.22} & \hm{98.33} \\
Base + TTC
& \hm{65.56} & \hm{98.61} & \hm{91.11} & \hm{6.94}
& \hm{90.74} & \hm{99.74} & \hm{91.11} & \hm{82.67}
& \hm{42.78} & \hm{100} & \hm{15.83} & \hm{12.50}  \\
FT
& \hm{68.15} & \hm{100.00} & \hm{4.72} & \hm{99.72}
& \hm{72.96} & \hm{99.72} & \hm{48.89} & \hm{70.28}
& \hm{38.98} & \hm{99.72} & \hm{8.06} & \hm{9.17} \\
FT + TTC
& \hm{66.67} & \hm{100} & \hm{0.00} & \hm{100}
& \hm{33.33} & \hm{100} & \hm{0.00} & \hm{0.00}
& \hm{33.33} & \hm{100} & \hm{0.00} & \hm{0.00} \\
GED-FT
& \hm{89.07} & \hm{98.89} & \hm{97.22} & \hm{71.11}
& \hm{98.33} & \hm{98.61} & \hm{97.50} & \hm{98.89}
& \hm{97.87} & \hm{98.61} & \hm{96.94} & \hm{98.06}  \\
GED-FT + TTC
& \hm{93.61} & \hm{100} & \hm{85.56} & \hm{94.28}
& \hm{94.35} & \hm{100} & \hm{92.78} & \hm{89.17}
& \hm{87.13} & \hm{100} & \hm{75.28} & \hm{85.11} \\
\midrule

\multicolumn{13}{l}{\textbf{Perturbed}} \\
Base
& \hm{67.87} & \hm{56.39} & \hm{90.28} & \hm{56.94}
& \hm{63.43} & \hm{14.72} & \hm{91.39} & \hm{84.17}
& \hm{19.81} & \hm{14.72} & \hm{23.06} & \hm{21.67}  \\
Base + TTC
& \hm{25.00} & \hm{14.72} & \hm{58.33} & \hm{1.94}
& \hm{38.06} & \hm{11.94} & \hm{58.33} & \hm{43.89}
& \hm{33.89} & \hm{98.89} & \hm{3.06} & \hm{0.00} \\
FT
& \hm{67.13} & \hm{100.00} & \hm{2.22} & \hm{99.17}
& \hm{61.02} & \hm{97.50} & \hm{34.72} & \hm{50.83}
& \hm{33.70} & \hm{97.22} & \hm{2.22} & \hm{1.67} \\
FT + TTC
& \hm{66.67} & \hm{100} & \hm{0.00} & \hm{100}
& \hm{33.33} & \hm{100} & \hm{0.00} & \hm{0.00}
& \hm{33.33} & \hm{100} & \hm{0.00} & \hm{0.00} \\
GED-FT
& \hm{96.39} & \hm{99.44} & \hm{95.83} & \hm{93.89}
& \hm{96.76} & \hm{93.61} & \hm{96.94} & \hm{99.72}
& \hm{93.70} & \hm{93.61} & \hm{91.39} & \hm{96.11} \\
GED-FT + TTC
& \hm{91.31} & \hm{100} & \hm{76.44} & \hm{97.50}
& \hm{95.93} & \hm{100} & \hm{91.39} & \hm{98.39}
& \hm{85.11} & \hm{100} & \hm{78.39} & \hm{77.50} \\
\bottomrule
\end{tabular}
\end{table*}

\subsubsection{Template Sensitivity}
Base models perform well on canonical traces but often reject admissible perturbed traces, even when every transition is accepted by the deterministic checker. This is clearest for T1 examples, where the correct judgment is unambiguous: the final answer is correct, the process is valid, and $Q_3=-1$. On baseline T1 traces, base-model $Q_2$ accuracies are 98.33\%, 93.89\%, and 92.22\% for GPT-OSS 20B, Qwen3-14B, and Phi-4-Reasoning. On perturbed T1 traces, these drop to 24.44\%, 16.11\%, and 14.72\%, respectively.
This gap suggests that the models are not only evaluating algebraic validity. They often treat the first unfamiliar transformation as an error, effectively mistaking procedural variation for incorrect reasoning. We refer to this behavior as \emph{template sensitivity}: the evaluator appears calibrated to common solution forms rather than to the local validity of adjacent steps. This matters beyond the specific models tested here. In any setting where multiple valid procedures can lead to the same result, a verifier that rewards only familiar paths may suppress correct but non-standard reasoning.

\subsubsection{Outcome and Process Are Entangled}
T3 examples separate final-answer correctness from process correctness. This design tests whether the model can judge the outcome and the reasoning process independently.
Under perturbation, this separation weakens. On T3 traces, base-model $Q_1$ accuracy drops from 96.67\% to 67.78\% for GPT-OSS 20B, from 71.39\% to 63.06\% for Qwen3-14B, and from 99.44\% to 56.94\% for Phi-4-Reasoning. Since the final answer remains correct by construction, these drops indicate that an unfamiliar or suspicious-looking derivation can contaminate final-answer judgment. Conversely, because the final answer is correct, models may also overlook the invalid intermediate step. This coupling is undesirable for automated evaluation: a grader should not reject a correct answer merely because the derivation is unusual, nor accept flawed reasoning merely because it reaches the right endpoint.

\subsubsection{Localization Remains the Bottleneck}
Error localization is the most fragile part of step-level verification. Detecting that a trace is invalid is easier than identifying the first invalid transition, especially when the trace contains valid detours mixed with actual errors. On perturbed traces, base-model $Q_3$ accuracy falls to 24.07\% for GPT-OSS 20B, 14.35\% for Qwen3-14B, and 19.81\% for Phi-4-Reasoning.
This limitation is important because localization is what turns a judgment into useful feedback. In tutoring, a model that says a solution is wrong but points to the wrong step can mislead the student. In verifier-guided inference, poor localization may cause the generator to revise the wrong part of its reasoning. More generally, process-level evaluation should not stop at binary correctness: reliable evaluators must also identify where a reasoning chain first becomes invalid.

\subsection{Impact of Supervised Fine-Tuning and Distillation}
Supervised adaptation improves robustness to perturbed traces, but the effect is model dependent. The results suggest that fine-tuning does not simply improve ``mathematical ability'' in a uniform way; instead, it shifts the model's decision boundary between accepting valid procedural variation and rejecting suspicious-looking traces.

For \textbf{GPT-OSS 20B}, GED-FT improves perturbed $Q_2$ from 54.06\% to 76.94\% and $Q_3$ from 24.07\% to 58.98\%. The strongest gain is on valid T1 perturbations, where GED-FT raises $Q_2$ from 24.44\% to 94.72\% and $Q_3$ from 24.72\% to 93.61\%. This indicates that distillation helps the model accept non-canonical but valid derivations, although some baseline trade-offs remain.

For \textbf{Qwen3-14B}, the best perturbed performance comes from FT+TTC, which raises localization from 14.35\% to 81.67\%. In contrast, GED-FT underperforms for this model, showing that teacher-rationale distillation is not uniformly beneficial. The usefulness of a distilled explanation appears to depend on the student model and on how well the explanation style matches its learned reasoning behavior.

For \textbf{Phi-4-Reasoning}, GED-FT is most effective, reaching 96.39\% on perturbed $Q_1$, 96.76\% on $Q_2$, and 93.70\% on $Q_3$. Standard FT, however, tends to over-accept traces, performing well on T1 while failing on T2. This shows why valid and invalid traces must be balanced: high acceptance of correct examples is not sufficient if the verifier loses its ability to reject flawed reasoning.

Overall, perturbation-oriented supervision can move models toward process verification, but it must be calibrated carefully. Robust verifier training should preserve canonical competence while also teaching models to accept valid alternatives and reject subtle invalid transitions.

\subsection{Test-Time Compute by Majority Voting}
Majority-vote TTC helps only when model errors are noisy rather than systematic. If a model has learned the relevant verification rule but occasionally samples an incorrect judgment, voting can reduce variance. If the model applies the wrong heuristic consistently, voting reinforces that heuristic.

The base-model results show this risk. For GPT-OSS 20B, TTC reduces perturbed $Q_1$ from 76.30\% to 32.04\% and $Q_2$ from 54.06\% to 31.11\%. Qwen3-14B and Phi-4-Reasoning show similar drops in perturbed $Q_1$. This suggests that more samples do not necessarily produce better verification; they can simply make an incorrect judgment more stable.

TTC is more useful after suitable adaptation. For Qwen3-14B, FT+TTC gives the strongest perturbed localization result, reaching 81.67\% on $Q_3$. In this case, training appears to provide a better verification policy, and TTC reduces residual uncertainty. Thus, it should be viewed as a calibration tool, not a substitute for learning trace-validity semantics.

\subsection{Limitations and Future Work}
This study is intentionally limited to single-variable linear equations with unique rational solutions. This narrow setting enables exact label validation and unambiguous first-error indices, but it does not establish that the same behavior holds for inequalities, systems of equations, proofs, programs, or scientific workflows. The purpose of this benchmark is therefore diagnostic: it isolates process-level verification in a setting where the ground truth can be controlled precisely.

Future work should extend the same procedural-evaluation framework to richer domains. In mathematics, this includes equations with branching conditions, multi-solution problems, and proof-style arguments. Beyond mathematics, the same idea may apply to software debugging, scientific protocols, chemical synthesis plans, and other workflows where the final output is not enough and the validity of intermediate operations matters. 
Our current analysis also motivates finer-grained studies in future. Per-perturbation-family breakdowns and held-out perturbation tests would help determine whether models learn general trace validity or adapt to specific transformation templates. We also report single-run adaptation results, so small differences between variants should be interpreted cautiously. Finally, Gemini-enhanced distillation depends on the teacher model, prompt design, and filtering policy; future work should evaluate whether the observed gains hold across different teachers and larger verified rationale sets.

\section{Conclusion}
We introduced a controlled benchmark for evaluating LLMs as step-level verifiers rather than only as problem solvers. In a linear-equation setting with exact trace labels, base models that perform well on canonical solutions often fail on admissible non-canonical traces, especially when asked to localize the first invalid step. These failures show that evaluator robustness is a distinct capability from final-answer accuracy.

The broader implication is that trustworthy LLM evaluation requires attention to process, not only outcomes. Many real tasks allow more than one valid path, and an evaluator should be able to accept legitimate procedural variation. Our results show that adaptation can reduce this robustness gap, but its benefits are model dependent, and majority-vote TTC can amplify systematic mistakes. Controlled process-level benchmarks therefore provide a potential tool for measuring and improving LLM reliability in domains where the reasoning path itself must be verified, such as automated software verification, legal reasoning, and clinical diagnostics.

\section*{Acknowledgment}
This work was supported in part by the National Science Foundation under Award No. 2434704.

\balance

\bibliographystyle{IEEEtran}
\bibliography{references}

@article{DBLP:journals/corr/abs-2110-14168,
  author = {Karl Cobbe and Vineet Kosaraju and Mohammad Bavarian and Mark Chen and Heewoo Jun and Lukasz Kaiser and Matthias Plappert and Jerry Tworek and Jacob Hilton and Reiichiro Nakano and Christopher Hesse and John Schulman},
  title = {Training Verifiers to Solve Math Word Problems},
  journal = {CoRR},
  volume = {abs/2110.14168},
  year = {2021},
  url = {https://arxiv.org/abs/2110.14168}
}

@article{DBLP:journals/corr/abs-2103-03874,
  author = {Dan Hendrycks and Collin Burns and Saurav Kadavath and Akul Arora and Steven Basart and Eric Tang and Dawn Song and Jacob Steinhardt},
  title = {Measuring Mathematical Problem Solving with the {MATH} Dataset},
  journal = {CoRR},
  volume = {abs/2103.03874},
  year = {2021},
  url = {https://arxiv.org/abs/2103.03874}
}

@techreport{OpenAI2025GPTOSS,
  author = {{OpenAI}},
  title = {GPT-OSS: Open-Weight Reasoning Models},
  institution = {OpenAI},
  year = {2025},
  url = {https://github.com/openai/gpt-oss}
}

@techreport{Alibaba2025Qwen3,
  author = {{Qwen Team}},
  title = {Qwen3 Technical Report},
  institution = {Alibaba Group},
  year = {2025},
  url = {https://github.com/QwenLM/Qwen3}
}

@techreport{Microsoft2025Phi4,
  author = {{Microsoft}},
  title = {Phi-4-Reasoning: Redefining Small Language Model Logic},
  institution = {Microsoft Research},
  year = {2025},
  url = {https://huggingface.co/microsoft/phi-4-reasoning}
}

@inproceedings{lightman2023lets,
  author = {Hunter Lightman and Vineet Kosaraju and Yuri Burda and Harrison Edwards and Bowen Baker and Teddy Lee and Jan Leike and John Schulman and Ilya Sutskever and Karl Cobbe},
  title = {Let's Verify Step by Step},
  booktitle = {International Conference on Learning Representations},
  year = {2024},
  url = {https://openreview.net/forum?id=v8L0pN6EOi}
}

@inproceedings{wang2024mathshepherd,
  author = {Peiyi Wang and Lei Li and Zhihong Shao and R. X. Xu and Damai Dai and Yifei Li and Deli Chen and Yu Wu and Zhifang Sui},
  title = {Math-Shepherd: Verify and Reinforce {LLMs} Step-by-Step without Human Annotations},
  booktitle = {Proceedings of the 62nd Annual Meeting of the Association for Computational Linguistics},
  year = {2024},
  pages = {9426--9439},
  url = {https://aclanthology.org/2024.acl-long.510/}
}

@inproceedings{chang2025hybridtts,
  author = {Kaiyan Chang and Y. Shi and C. Wang and H. Zhou and C. Hu and X. Liu and Y. Luo and Y. Ge and T. Xiao and J. Zhu},
  title = {Step-Level Verifier-Guided Hybrid Test-Time Scaling for Large Language Models},
  booktitle = {Proceedings of the 2025 Conference on Empirical Methods in Natural Language Processing},
  year = {2025},
  pages = {18462--18477}
}

@article{kamoi2025formalverifiers,
  author = {Ryo Kamoi and Yusen Zhang and Nan Zhang and S. S. Das Sarkar and Rui Zhang},
  title = {Generalizable Process Reward Models via Formally Verified Training Data},
  journal = {arXiv preprint arXiv:2505.15960},
  year = {2025}
}

@article{li2025verifybench,
  author = {X. Li and X. Li and S. Hu and Y. Guo and W. Zhang},
  title = {VerifyBench: A Systematic Benchmark for Evaluating Reasoning Verifiers across Domains},
  journal = {arXiv preprint arXiv:2507.09884},
  year = {2025}
}

@inproceedings{li2024gsm,
  author = {Qintong Li and Leyang Cui and Xueliang Zhao and Lingpeng Kong and Wei Bi},
  title = {{GSM}-Plus: A Comprehensive Benchmark for Evaluating the Robustness of {LLMs} as Mathematical Problem Solvers},
  booktitle = {Proceedings of the 62nd Annual Meeting of the Association for Computational Linguistics},
  year = {2024},
  pages = {2961--2984},
  url = {https://aclanthology.org/2024.acl-long.163/}
}

@article{mirzadeh2024gsmsymbolic,
  author = {Iman Mirzadeh and Keivan Alizadeh and Hanie Sedghi and Samy Bengio and Mehrdad Farajtabar},
  title = {{GSM}-Symbolic: Understanding the Limitations of Mathematical Reasoning in Large Language Models},
  journal = {arXiv preprint arXiv:2410.05229},
  year = {2024},
  url = {https://arxiv.org/abs/2410.05229}
}

@article{huang2025mathperturb,
  author = {K. Huang and J. Guo and Z. Li and X. Ji and J. Ge and W. Li and Y. Guo and T. Cai and H. Yuan and R. Wang and Y. Wu and M. Yin and S. Tang and Y. Huang and C. Jin and X. Chen and C. Zhang and M. Wang},
  title = {{MATH}-Perturb: Benchmarking {LLMs}' Math Reasoning Abilities against Hard Perturbations},
  journal = {arXiv preprint arXiv:2502.06453},
  year = {2025},
  url = {https://arxiv.org/abs/2502.06453}
}

@article{chatziveroglou2025exploring,
  author = {George Chatziveroglou and Ryan Yun and Michael Kelleher},
  title = {Exploring {LLM} Reasoning through Controlled Prompt Variations},
  journal = {arXiv preprint arXiv:2504.02111},
  year = {2025},
  url = {https://arxiv.org/abs/2504.02111}
}

@article{numericalsensitivity2025,
  author = {Z. Sun and G. Dai and I. Tsang and H. Ye},
  title = {Numerical Sensitivity and Robustness: Exploring the Flaws of Mathematical Reasoning in Large Language Models},
  journal = {arXiv preprint arXiv:2511.08022},
  year = {2025},
  url = {https://arxiv.org/abs/2511.08022}
}

@inproceedings{BhandariPardosEDM2025Autograder,
  author = {S. Bhandari and Z. Pardos},
  title = {Can Language Models Grade Algebra Worked Solutions? {Evaluating} {LLM}-Based Autograders against Human Grading},
  booktitle = {Proceedings of the 18th International Conference on Educational Data Mining},
  year = {2025},
  pages = {554--558}
}

@inproceedings{chen2024teval,
  author = {Z. Chen and W. Du and W. Zhang and K. Liu and J. Liu and M. Zheng and J. Zhuo and S. Zhang and D. Lin and K. Chen and F. Zhao},
  title = {{T-Eval}: Evaluating the Tool Utilization Capability of Large Language Models Step by Step},
  booktitle = {Proceedings of the 62nd Annual Meeting of the Association for Computational Linguistics},
  year = {2024},
  pages = {9510--9529}
}

@misc{Google2025Gemini3,
  author       = {{Google}},
  title        = {Introducing {Gemini 3 Flash}: Benchmarks, Global Availability},
  year         = {2025},
  note         = {[Online]. Available: \url{https://blog.google/products-and-platforms/products/gemini/gemini-3-flash/}}
}

@inproceedings{mazdarani2026,
  author    = {Fateme Mazdarani and Alberto Campos Hernandez and Carlos Toxtli},
  title     = {Worker-Centered {AI}: Transparent Explanations for Trustworthy Task Recommendation in Crowd Work},
  booktitle = {2026 IEEE Conference on Artificial Intelligence},
  year      = {2026},
  pages     = {295--302},
  doi       = {10.1109/CAI68641.2026.11536153}
}
\end{document}